\documentclass[11pt,a4paper]{article}
\usepackage[hyperref]{emnlp2020}
\usepackage{times}
\usepackage{latexsym}
\usepackage{numprint}
\usepackage{multirow}

\usepackage{microtype}
\usepackage{graphicx}
\usepackage{amsmath}
\usepackage{booktabs}
\aclfinalcopy 

\title{Generating Dialogue Responses from a Semantic Latent Space}

\author{Wei-Jen Ko$^1$\ \ \ \ 
Avik Ray$^2$\ \ \ \ 
Yilin Shen$^2$\ \ \ \ 
Hongxia Jin$^2$\\
$^1$ The University of Texas at Austin\\ $^2$ Samsung Research America\\

{\small 
{\tt wjko@utexas.edu,} \tt\{avik.r, yilin.shen, hongxia.jin\}@samsung.com}}

\date{}

\begin{document}
\maketitle
\begin{abstract}
Existing open-domain dialogue generation models are usually trained to mimic the gold response in the training set using cross-entropy loss on the vocabulary. However, a good response does not need to resemble the gold response, since there are multiple possible responses to a given prompt. In this work, we hypothesize that the current models are unable to integrate information from multiple semantically similar valid responses of a prompt, resulting in the generation of generic and uninformative responses. To address this issue, we propose an alternative to the end-to-end classification on vocabulary. We learn the pair relationship between the prompts and responses as a regression task on a latent space instead. In our novel dialog generation model, the representations of semantically related sentences are close to each other on the latent space. Human evaluation showed that learning the task on a continuous space can generate responses that are both relevant and informative.
\end{abstract}

\section{Introduction}

The sequence-to-sequence framework and transformer based models are the most popular choices for designing open-domain neural response generation systems \cite{ncm,tt}.
Those models typically involve maximizing the probability of the ground truth response given the input prompt, trained using a cross entropy loss on the vocabulary. However, dialogue response generation is an open-ended, high entropy task, since there can be a wide variety of possible responses to a given prompt. A good response does not have to use similar vocabulary or similar sentence structure as the gold response, thus the end-to-end cross entropy loss is unsuitable for this task. We hypothesize that this fundamental deficiency is the primary reason why dialog generation models tend to generate bland and uninformative responses, such as {\em ``I don't know"} \cite{serban}, despite the presence of much more instances of specific responses in the training data than generic responses.



The specific issue is the following. A model trained using maximum likelihood objective treats each token of the vocabulary independently. The probabilities of each individual informative word in the vocabulary are low because the answer is open-ended. The model is unable to capture that most of the probability mass are on a group of semantically related words.
Thus the words with the highest probabilities are often uninformative stop words with high frequency in the training data. A similar effect happens on the utterance level when using beam search decoding. When searching for the most probable utterance, the probability of each candidate sentence is calculated independently, and the model is unable to use the semantic relatedness between different candidate utterances \cite{mr}. While informative and specific responses collectively have a high probability, it is diluted by the large number of variations and possibilities of specific responses. On the other hand, generic responses have much less variations, thus they become the most probable response sequences. An alternative decoding method to beam search is sampling \cite{ns,ts}, which does not suffer from this problem. However, sampling does not consider the subsequent words during decoding, and the randomness in word choice makes it prone to generating implausible responses, responses with grammatical errors and coherence issues.

Aiming to take into account the semantic relatedness of diverse specific responses, we propose an alternative to cross-entropy training, which is learning the pair relationship between the prompts and responses as a regression task on a latent space.  

In our novel dialog generation model, the generation process could be separated into two steps. The first step predicts a sentence vector of the response on the latent space, on which the representations of semantically related sentences are close to each other. Since predicting a vector is a regression problem in the latent space instead of classification in the vocabulary as in MLE loss, our model is able to learn that most of the probability mass of the response is around the cluster of possible specific responses. 
This is illustrated in Figure \ref{fig:visualization} showing our model's representations of prompts and responses on a t-SNE plot.

The second step constructs the full response sentence from the predicted vector. We train an autoencoder. The decoder part is used for constructing the full response sentence from the predicted sentence vector.  
Since the semantics of the response and the decoding are learned separately, we can perform beam search for the most probable sequence given the semantic vector during inference without preferring generic responses. 

The main contributions of our work are 1) We propose to learn dialogue generation as an regression task on a semantic latent space, as an alternative of end-to-end cross entropy training used in most previous methods, to address the problem that end-to-end cross entropy classification  are unable to integrate information from semantically similar responses and words. 2) Our model separates the response into information likely and unlikely to be correlated with the prompt. 3) Evaluation by crowdworkers showed that the latent space method significantly outperforms baselines using end-to-end cross entropy classification, in terms of generating responses that are both relevant and informative.

\section{Related work}
Several previous models also use the idea of learning on sentence vector representations.  \citet{AEM} used two autoencoders to learn the semantic representations of inputs, and learned utterance-level dependency between those representations. Spacefusion\cite{SF} fuses the autoencoder and seq2seq feature space, so that the distance and direction from a predicted response vector roughly matches the relevance and diversity. Those methods add additional autoencoder losses to manipulate the intermediate representation space, but they still use the problematic end-to-end cross entropy loss for generation. In our work, we completely remove the end-to-end loss term, so the matching of the input and response is learned only on a shared semantic latent space. \citet{mr} proposed a two-stage generation process, which predicts the average of the reference responses as an intermediate task, but it requires multiple responses for each prompt in the training data. On the machine translation task, \citet{VMF} explored predicting continuous vectors on the word level in seq2seq models instead of using softmax classification. We predict continuous vectors on the utterance level. 

There are other aspects to tackle the generic response problem, \citet{mmi} maximized mutual information in decoding or reranking. \citet{zhou} trained multiple response mechanisms to model diversity. \citet{shao} split the generation into segments and allow attention to attend to both the prompt and the response to improve diversity. Several works use explicit specificity metrics to manipulate the specificity of the responses. Frequency based metrics such as IDF are used in \cite{DR,zhang}. \citet{ko} proposed using a specificity metric trained on discourse relation pair data.

We use a modified version of deep canonical correlation analysis (DCCA)\cite{DCCA} to learn the semantic latent space. DCCA has previously been used on various tasks including feature learning \cite{DCCAE}, caption retrieval \cite{DCMIT}, multi-label classification \cite{C2AE}, image  cross-reconstruction \cite{corrnn}, and multilingual word similarity \cite{DPCCA}. \cite{dcclstm} experimented on performing DCCA on sequential data with a recurrent network.

\begin{figure*}
\begin{center}
\includegraphics[width=0.9\textwidth]{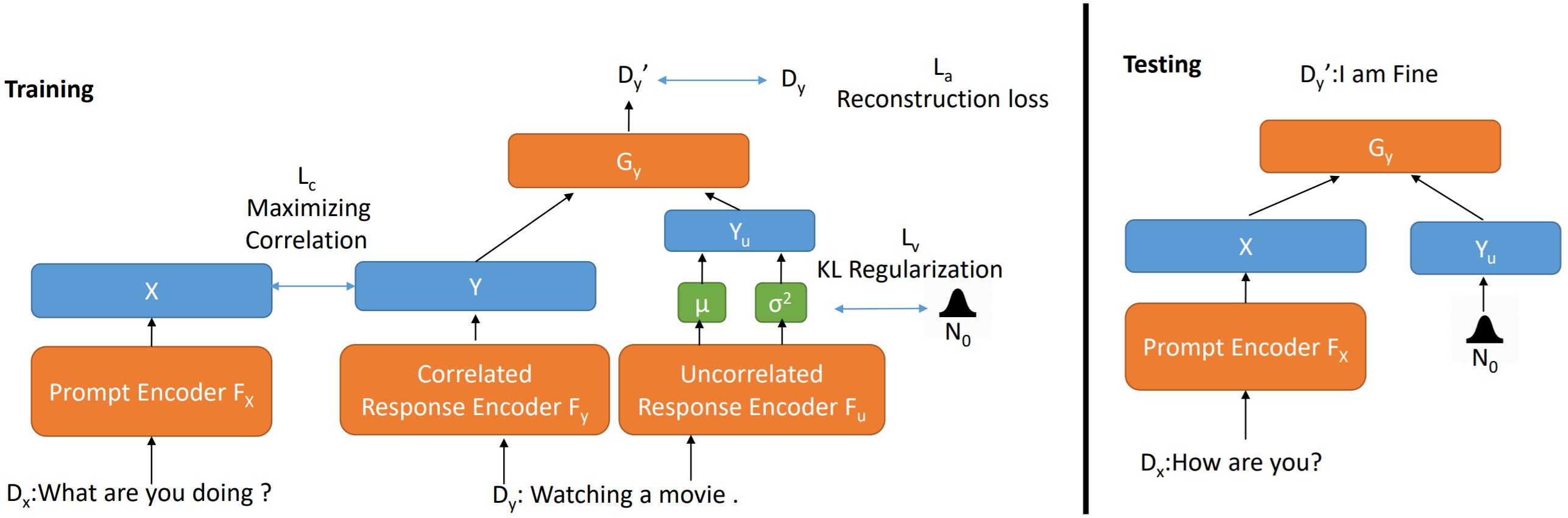}
\end{center}
\caption{Model Architecture. We learn the pair relation between prompts $D_x$ and responses $D_y$ as an regression task on the shared semantic latent space of $X$\&$Y$, so the model is able to aggregate information from semantically similar sentences. We train an autoecoder to construct the response sentence from the latent space. An uncorrelated representation $Y_u$ is allowed in the autoencoder to represent information unlikely to be related to the prompt.}
\label{fig:model}
\end{figure*}

\section{Our Method}
Given example dialogue (Prompt, Response) pairs $(D_{x},D_{y})$ from open-domain dialogue datasets, our goal is to generate a relevant and non-generic response when given an unseen prompt. The structure of our model is depicted in Figure \ref{fig:model}. It consists of three encoders; {\em Prompt Encoder} $F_x,$ {\em Correlated Response Encoder} $F_y,$ and {\em Uncorrelated Response Encoder} $F_u.$ The final response is generated from a semantic latent vector via a {\em Decoder} $G_y.$ During training all three encoders and the decoder are tuned. However, during testing only the Prompt Encoder and the Decoder are utilized. 

\subsection{Learning the correlated semantic latent space} \label{sec:latent_space}

Our aim is to learn a latent space where representations of semantically related prompts, encoded by $F_x$, are close to each other, and so are semantically related responses encoded by $F_y$. Furthermore, we want a prompt encoded by $F_x$ to be close to its corresponding responses encoded by $F_y$.

For this we employ canonical correlation analysis (CCA) \cite{cca} between the prompt and response pairs. We maximize the correlation of the embeddings with the other sentence in the pair. Since semantically similar responses are likely to correspond to a similar set of prompts, semantically similar sentences will have similar representations in the CCA encoded space. Generic responses could be responses to a much larger set of prompts, so they will have very different representations in the latent space, thus they could be separated with specific responses.

We use two recurrent neural networks $F_{x},F_{y}$ as feature extractors to map prompts and responses into the shared featured space respectively. Using the definition of CCA, we maximize the total correlation of each dimension between $X=F_{x}(D_{x})$, $Y=F_{y}(D_{y})$ as follows.

\begin{equation}
\max\left(\sum\limits_{i=1}^{k}\mbox{corr}(X^{i},Y^{i})\right) =
\nonumber
\end{equation}
\vspace{-0.15in}
\begin{equation}
\max \left( \sum\limits_{i=1}^{k} \frac{\sum\limits_{m}(X^{i}_{m}-\bar{X^{i}})(Y^{i}_{m}-\bar{Y^{i}})}{\sqrt{\sum\limits_{m}(X^{i}_{m}-\bar{X^{i}})^2\sum\limits_{m}(Y^{i}_{m}-\bar{Y^{i}})^2}} \right), \label{eq:cca1}
\end{equation}

subject to the condition
\begin{equation}
\forall\{i,j|i\neq j\}: \sum\limits_{m}(X^{i}_{m}-\bar{X^{i}})(X^{j}_{m}-\bar{X^{j}})=
\nonumber
\end{equation}
\vspace{-0.15in}
\begin{equation}
\sum\limits_{m}(Y^{i}_{m}-\bar{Y^{i}})(Y^{j}_{m}-\bar{Y^{j}})=0,
\label{eq:cca_condition}
\end{equation}
$i,j$ are the indices of the feature dimension. $\bar{X^{i}},\bar{Y^{i}}$ are the mean of the i-th feature dimension. $m$ is the index of the example pair in the batch.  $k$ is the number of feature dimensions. The condition ensures that the different dimensions in the representation are uncorrelated, to avoid redundant representations.

To make the two feature spaces $X$ and $Y$ shared, we add the following conditions to the mean and variance of both representations, inspired by \cite{C2AE}. 

\begin{equation}
\forall i: \bar{X^{i}}=\bar{Y^{i}}=0 \label{eq:cca_deriv1}
\end{equation}
\begin{equation}
\forall i: \sum\limits_{m}(X^{i}_{m})^2=\sum\limits_{m}(Y^{i}_{m})^2=C
\end{equation}
, where C is an arbitrary constant, we use C=1.

When prompt $X$ and response $Y$ are perfectly correlated, and these two conditions perfectly hold, $X$ will be equal to $Y$, so this makes $X$ and $Y$ interchangeable during inference. This is desirable because we do not have access to Y during inference.

Using the two conditions, Equations \ref{eq:cca1}, \ref{eq:cca_condition} becomes\footnote{ In practice, the correlation is calculated for each batch separately, so it is important that the process of dividing training data into batches is random, and the batch size is sufficiently large.}:
\begin{equation}
\max \left(\sum\limits_{i=1}^{k}corr(X^{i},Y^{i})\right)
\nonumber
\end{equation}
\vspace{-0.15in}
\begin{equation}
= \max\left(\sum\limits_{i=1}^{k} \sum\limits_{m}X^{i}_{m}Y^{i}_{m} \right)
\nonumber
\end{equation}
\vspace{-0.15in}
\begin{equation}
= \min \left( \sum\limits_{i=1}^{k} \sum\limits_{m}(X^{i}_{m}-Y^{i}_{m})^2 \right),
\end{equation}
subject to,
\begin{equation}
\sum\limits_{m}X^{i}_{m}X^{j}_{m}=\sum\limits_{m}Y^{i}_{m}Y^{j}_{m}=0 \label{eq:cca_deriv4}
\end{equation}
We can formulate the total CCA loss from \eqref{eq:cca_deriv1} to \eqref{eq:cca_deriv4} as,
\begin{equation}
{L}_{c}=\sum\limits_{i=1}^{k}  \left(\sum\limits_{m}(X^{i}_{m}-Y^{i}_{m})^2+ \right.
\nonumber
\end{equation}
\vspace{-0.15in}
\begin{equation}
\lambda_{1}\left(\lvert\sum\limits_{m}X^{i}_{m}\rvert+\lvert\sum\limits_{m}Y^{i}_{m}\rvert\right)+
\nonumber
\end{equation}
\vspace{-0.15in}
\begin{equation}
\left. \lambda_{2}\left(\lvert\sum\limits_{m}(X^{i}_{m})^2-1\rvert+\lvert\sum\limits_{m}(Y^{i}_{m})^2-1\rvert\right)\right) 
\nonumber
\end{equation}
\vspace{-0.15in}
\begin{equation}
 +\lambda_{3}\sum\limits_{i,j}^{i\neq j}\left(\lvert\sum\limits_{m}X^{i}_{m}X^{j}_{m}\rvert+\lvert\sum\limits_{m}Y^{i}_{m}Y^{j}_{m}\rvert\right)
\end{equation}
where $\lambda_{1},\lambda_{2},\lambda_{3}$ are tunable hyper-parameters.
\subsection{Generating the response from the semantic latent space} \label{sec:generation}
Since $X$ and $Y$ are interchangeable in the semantic space, we directly use the features extracted from the prompt $X$ to approximate the response features $Y.$  Now we want to generate the response sentence $D_{y}'$ from the latent space representations. For this purpose, an additional autoencoder is trained on all the training set responses, simultaneously with the CCA. The autoencoder consists of encoder $F_{y},$ and decoder $G_{y},$  both of which are recurrent networks. The parameters of the encoder are shared with the semantic feature extractor of the responses. During inference, features extracted from the prompt $X$ are directly fed into the decoder to generate the response sentence.\footnote{We also experimented on adding domain discriminative adversarial training \cite{ADDA} between $X$ and $Y$, but it did not improve the results. This shows that our conditions (1),(3),(4) already make the distribution of the two encoders sufficiently similar.}

\begin{equation}
D_{y}'=G_{y}(F_{x}(D_{x}))
\end{equation}

Generating sentences from a continuous space is known to produce ungrammatical text \cite{gscs}. To address this issue, we replace some autoencoder input word tokens, by the unknown word token $\langle unk \rangle$. The probability each word is chosen to be replaced is independent and uniform. The replacing serves three purposes. First, it makes the decoder more robust, and able to generate grammatical responses when there is noise in the decoder input. This is important because the decoder input during training and inference are from different encoders. Second, it prevents the autoencoder from overfitting too early before the CCA objective converges. Finally, masked language models have been shown successful on learning representations of sentences suitable for a wide range of tasks \cite{bert}. This is desirable since this representation is also used for learning the CCA for the semantic latent space.

\subsection{Correlated and uncorrelated representations} \label{sec:correlated_uncorrelated}

When encoding the response with $F_{y}$, the autoencoder and the CCA loss have conflicting objectives. The autoencoder task requires the representation to preserve all the information in the sentence for reconstruction. The CCA task aims to preserve only the information likely to be related to the prompt, and discard all other irrelevant information. 
For example, a paraphrase pair are likely to be valid responses to the same prompts, so they should have the same representation under CCA objective, but the autoencoder objective forces the representations to be different, to enable reconstruction of the exact sentences. A response could also include a topic change, which makes part of the response completely irrelevant to the prompt, and that information should not be in the CCA representation.

To model this issue, we separate the autoencoder representations into the correlated part $Y$, which correlates with the prompt, and the uncorrelated part $Y_{u}$. The correlated part learns both the autoencoder task and the CCA task. The uncorrelated part is only trained for autoencoder reconstruction.

During training, $G_{y}$ learns to reconstruct from the concatenation of the correlated and uncorrelated representations. The reconstruction is trained using cross entropy loss.
\begin{equation}
D_{y}'=G_{y}([Y;Y_u])=G_{y}([F_{y}(D_{y});F_{u}(D_{y})])
\end{equation}
\begin{equation}
L_{a}=\mbox{Cross Entropy}(D_y,D_y')
\end{equation}

During testing, $G_{y}$ generates the response from the CCA semantic representation of the prompt and a vector $R$ representing the uncorrelated part of the response.
\begin{equation}
D_{y}'=G_{y}([F_{x}(D_{x});R])
\end{equation}

By adding additional regularization to the uncorrelated representation $Y_u$ during training, we encourage a normal distribution with zero mean and unit variance for each dimension. Hence during inference we can sample $R$ from this distribution or use a fixed prior to approximate $Y_u.$ 

The formulation of the regularization is the same as variational autoencoders \cite{VAE}. An encoder recurrent network $F_{u}$ predicts a mean $\mu$ and variance $\sigma^2$ for each dimension, and $Y_{u}$ is sampled from that multivariate normal distribution. The predicted $\mu$ and $\sigma^2$ is regularized by the KL divergence with unit normal distribution.
\begin{equation}
L_{v}=\sum\limits_{i}\sum\limits_{m}((\mu^i_m)^2+(\sigma^i_m)^2-log((\sigma^i_m)^2))
\end{equation}

With this regularization, $R$ can either be set to all zeroes , or be randomly drawn from a unit normal distribution. We found that the generated sentence is insensitive to this choice, so the two ways generate exactly the same sentence more often than not. Despite the insensitivity, we calculated the ratio between the KL loss and the autoencoder reconstruction loss, and found that the ratio consistently increases during training, indicating that there is no posterior collapsing \cite{VLA}. We also found that adding the uncorrelated representation allows both the CCA loss and the autoencoder reconstruction loss to converge to a significantly lower value. Inspection on generated sentences showed that there is obvious improvement on relevance, at the cost of slightly more frequent grammatical errors.\footnote{To avoid introducing excessive noise while using $R$ as an approximation, we use only 10 dimensions for the uncorrelated part. Higher number of dimensions resulted in worse performance in our experiments.}


During training, the gradients of all loss terms are weighted and summed and all parameters are updated together. The total loss is:
\begin{equation}
\mathcal{L}=\lambda_4L_{c} +\lambda_5L_{a} +\lambda_6L_{v}    
\end{equation}
where $\lambda_{4},\lambda_{5},\lambda_{6}$ are hyper-parameters.

\subsection{Attention} \label{sec:attention}
The described model does not have an attention mechanism, so it cannot dynamically focus on different parts of the prompt during generation. We also experiment with a variant of our model with attention \cite{Luong}. Similar to previous works, the key and value is from the RNN hidden state of the prompt encoder $F_x$, and the query is the hidden state of the response decoder $G_y$. To prevent nullifying the main purpose of our model design: removing end-to-end MLE training, we create a bottleneck to limit the end-to-end information flow before concatenating the attention output vector with the hidden state. The bottleneck is a fully connected layer that reduces the attention output vector into a low dimension.\footnote{We use dimension 10. Without the bottleneck, the model will only rely on attention and completely ignore $F_x$ and $F_y$, effectively degenerating into a Seq2Seq model. }

\begin{table*}
\centering
\small
\setlength{\tabcolsep}{0.5em}
\begin{tabular}{c|l|lllll|lll}
  & &\multicolumn{5}{c|}{Automatic metrics} &\multicolumn{3}{c}{Human evaluation}   \\ \toprule
  Dataset& Model & Bleu-1 & Bleu-2 & Sim & Dist-1& Dist-2&Rel.&Info.&UI\\
  \midrule
  \multirow{8}{*}{PersonaChat} & MLE+beam search& 0.146 &0.0640& 0.854 &0.0128&0.040&1.09&0.98&1.43\\
  & MLE+sampling& 0.148&0.0605 & 0.851&0.0313&0.131&1.05&1.39&1.65 \\
    & MMI& 0.124&0.0551 & 0.838&\bf{0.0517}&0.217&1.18&1.08&1.54 \\
  &SpaceFusion & 0.176 & 0.0715 & 0.855 &0.0233 & 0.077 & 1.05&\bf{1.66}&1.84\\
  & Ours&\bf{0.191} &\bf{0.0746}&\bf{0.871}& 0.0363 & 0.179 &1.29&1.35&\bf{1.97} \\
  & Ours+Attention& 0.182&0.0712&0.868&0.0360  &0.171  &\bf{1.32}&1.22 &1.76\\
  & Ours w/o Uncorrelated&0.179 &0.0729&0.866&0.0305  &0.127&0.96&1.40&1.46\\
  &Ours w/o Denoising&0.167 &0.0556&0.868&0.0332  &\bf{0.252}&-&-&-\\
\midrule
  \multirow{6}{*}{DailyDialog}  & MLE+beam search& 0.100 &0.0394& 0.763 &0.0440&0.152&1.25&0.73&0.84\\
  & MLE+sampling&0.142&0.0413 & 0.790&0.0620&\bf{0.389}&0.81 &\bf{1.56}&1.18 \\
    & MMI& 0.094&0.0369 & 0.764&\bf{0.0697}&0.270&\bf{1.29}&0.89&1.09 \\
 
  &SpaceFusion & 0.146 & \bf{0.0595} &0.792 & 0.0531 & 0.216 & 1.21&1.08&1.24\\
  & Ours& 0.170 &0.0575& \bf{0.807} & 0.0457 &0.191 &1.18&1.45&1.71 \\
  & Ours+Attention&\bf{0.171}  &0.0558&0.805  &0.0530 &0.213& 1.24&1.51&\bf{1.74} \\

\bottomrule
\end{tabular}

\caption{Result comparison of all models on PersonaChat and DailyDialog datasets using different automatic and human evaluation metrics. Our model generates responses that are both informative and relevant. }
\label{tab:results}

\end{table*}
\section{Experiments}

\subsection{Methodology}
We conduct experiments on two datasets: PersonaChat \cite{pc} and DailyDialog \cite{dd}. PersonaChat is a chit-chat dataset collected by crowdsourcing. We do not use the personas in the dataset since they are not related to our work. We use \numprint{122499} prompt-response pairs for training, \numprint{3000} pairs for validation and \numprint{4801} pairs for testing. DailyDialog is a collection of conversations in daily life for English learners. We remove those prompt-response pairs in the validation and test set that also appears in the training set, which resulted in about $30\%$ of pairs removed in the test set. The final dataset has \numprint{76052} pairs for training, \numprint{5334} pairs for validation, and \numprint{4738} pairs for testing.

We compare our models with the vanilla Seq2seq model with attention \cite{Luong} trained using cross entropy loss, decoded using beam search and nucleus sampling \cite{ns}. We also compare with previous works MMI-anti \cite{mmi} and SpaceFusion \cite{SF}. MMI-anti also addresses the generic response issue. It is based on mutual information, and improves the Seq2seq model by penalizing frequent responses with an anti-language model. SpaceFusion is a recent method which learns a fused common space representation of the Seq2seq dialogue generation task and the autoencoder task, it falls on the same line of work that tries to manipulate the latent space representations. We use the authors' code. For these two baselines and our models, we report the result of the decoding method (beam search or sampling) that performed better. The goal of our experiments is to compare learning on the latent space with end-to-end cross entropy training.

We use $1$ layer GRU for all encoders and decoders. The correlated representation size is $512$, the uncorrelated representation size is $10$. We implement $F_{y}$ and $F_{u}$ as different output dimensions of the same GRU.
For compared methods we use hidden layer size $522$. The word embedding dimension is $128.$ 
 We use Adam optimizer with learning rate $0.001,$ $\beta_1=0.9,  \beta_2=0.999$. Batch size is $64$. $\{\lambda_1,\lambda_2,\lambda_3,\lambda_4,\lambda_5,\lambda_6 \}$ is set to $\{$3.9,6.25,0.05,2,2,0.1$\}$, they are tuned to make the conditions (2), (3), (4) enforced properly.  For our model with attention, the attention bottleneck has dimension $10.$ Our models were trained on one Tesla M40 GPU, and the run time was less than 2 hours.

\begin{table*}[t]

\small

\begin{tabular}{l|l}

 \toprule
 \multicolumn{2}{c}{PersonaChat}\\
  \midrule
{\bf Prompt}&~~wow i love to see traditional dance .\\
  \midrule
{\bf MLE+beam search}& ~~do you have a favorite color ?\\

{\bf MMI} &~~yes , i love the color blue . \\
{\bf Spacefusion} &~~i am not sure . i am a college student . \\
{\bf Ours} &~~its fun . i've always wanted to go for a dance class . \\
 \midrule
 {\bf Prompt}&~~do you go to school at all\\
 \midrule
{\bf MLE+beam search}& ~~i do not have a job but i want to be a doctor\\

{\bf MMI} &~~sometimes , but i do like photography . \\
{\bf Spacefusion} &~~i do not . i am a college student . \\
{\bf Ours} &~~i don't . i'm studying to go to college . \\
 \midrule
 {\bf Prompt}&~~i live in nebraska . i don't think i have ever been to ohio before .\\
  \midrule
{\bf MLE+beam search}& ~~oh wow . what do you do for a living ?\\

{\bf MMI} &~~ oh do you do volunteer work \\
{\bf Spacefusion} &~~oh wow . i love to go to the beach . \\
{\bf Ours} &~~wow , i have been there . do you have a favorite place ? \\
  \midrule

 \multicolumn{2}{c}{DailyDialog}\\
  \midrule

 {\bf Prompt}&~~Can you operate computers skillfully ?\\
  \midrule
{\bf MLE+beam search}& ~~Yes , I am .\\     

{\bf MMI} &~~Yes , I have a special job . \\
{\bf Spacefusion} &~~Yes , I can . I have any other Word 2003 , and I can live in other areas . \\
{\bf  Ours+Attention} &~~No , I am not familiar with both Java and C Programming Languages . \\
 \midrule
 {\bf Prompt}&~~Exporters must ensure that their product satisfies customers ' needs , wants and likes .\\
 \midrule
{\bf MLE+beam search}& ~~Do you have any other questions ?\\

{\bf MMI} &~~We have to buy a new set . \\
{\bf Spacefusion} &~~I am calling to see you . But I've got a new customer here . But I don't want to be\\&~~ aware of their house .\\ 
{\bf Ours+Attention} &~~That means that we have a commission about selling electronics . The prices are\\&~~ perfect for you to promote the commission and the prices are cheaper .\\
 \midrule
 {\bf Prompt}&~~It closes at 7:00 on Sundays .\\
  \midrule
{\bf MLE+beam search}& ~~You're welcome , I can help you .\\

{\bf MMI} &~~How do you get that ?\\
{\bf Spacefusion} &~~Nice to see you . The alarm isn't a fool . \\
{\bf  Ours+Attention} &~~Hold on , I have some other classes available . \\
 \bottomrule
 \end{tabular}

\caption{Example model responses}
\label{tab:example_responses}
\end{table*}

\subsection{Human evaluation}
Human evaluation is the only reliable way to evaluate this task. Following the standards of \cite{shao,rw,zhang,mr}, for each model, we randomly sample $500$ pairs from our test set, and each pair is rated by $3$ distinct crowdsourcing workers.\footnote{We use this crowdsourcing platform: https://appen.com/ } Our evaluation involved a pool of around 400 distinct workers in total. We ask them to rate 1) informativeness and specificity of each response and 2) if the response is on topic (relevance), on a Likert scale from $0$ to $3.$ 
We report the average of all annotations for both metrics. 
We also design a new metric, the useful information (UI) score, which is multiplying informativeness with relevance for each sentence, and then take the average over all sentences. The intuition of this score is that a response has to be both informative and on topic to be good. If one response is very informative, but not on topic, then all of the information it provides is useless. On the other hand, if a response is answering the prompt, but uninformative, it is a generic response which is undesirable. Thus the UI score could be used to approximate the quality of the responses.

We mix some quality control sentences with the model responses in the task for workers. Those workers who failed to rate these quality control sentences reasonably were excluded. We mix the responses from different models in the tasks given to each worker, so the bias of individual workers would not affect the relative performance between different models. The interannotator pearson correlation is $0.55$.

The results are shown in Table 1. Note that there is a trade-off between informativeness and relevance, since a generic response can reply to a wide range of prompts, it will be easier for them to be will, and informative responses are more specialized, thus it would be more difficult for them to be relevant. Example model responses are shown in Table 2.

On the PersonaChat dataset, SpaceFusion and MLE+sampling could generate very informative responses, but the relevance score is low, indicating the responses are often not on topic. Our model outperforms MLE+beam search and MMI on both relevance and informativeness. On the DailyDailog dataset, MLE+sampling scores highest on informativeness, but the responses are not on topic. Both MMI and MLE+beam search are relevant but prone to generic responses. Our models are the only ones that could be both informative and coherent. Adding attention to our model improves both relevance and informativeness on the DailyDialog dataset, but harms informativeness on the PersonaChat dataset. Our models performs best on the UI score for both datasets. We performed bootstrapping significance test, and found that our improvements are statistically significant. \subsection{Automatic evaluation}
Almost all existing automatic metrics for dialog generation compares the generated response and the gold response in some way. However, a good response could be open-ended and doesn't have to resemble the gold response. \citet{liu} showed that automatic metrics have low correlation with human judgements. Furthermore, because our model is not trained to mimic the gold response, these metrics are especially unsuitable for evaluating our model. Take perplexity for example, when training a vanilla Seq2seq model on PersonaChat, the test perplexity could achieve $\sim$ 38 \cite{pc}. For our model, the test perplexity is very high ($10^3\sim10^4$), because unlike previous methods, our model does not optimize for low cross-entropy loss on the vocabulary. Nonetheless, human evaluation obviously prefer our model over Seq2seq, verifying that low perplexity is not necessary for a good model. 
Despite the deficiencies of those automatic metrics, they are still widely used because there are no good alternatives. For reference only, we include the results of the following automatic metrics for reference in Table 1: (1) BLEU-1 and BLEU-2 \cite{bleu} (2) Embedding Average cosine similarity \cite{cossim} between the sentence vectors of the generated and gold response. The sentence embedding is computed by averaging the GloVe embedding of each word in the sentence. This metric measures the coherence of the response. (3) dist-1 and dist-2 \cite{mmi}, which evaluates the diversity of the generated responses.  They respectively calculate the count of distinct unigrams and bigrams, divided by the total number of words in all responses. Those metrics are also used in \cite{zhang,cohh,WAE,mr}. 
 
  For the BLEU scores, we can see that even though we do not train to mimic the gold responses, Our model still gets higher BLEU than most of the baselines, showing the effectiveness of our latent space method over MLE training. For the embedding similarity score, our model consistently outperform other compared methods. 
The calculation of Dist scores involves the sentence length in the denominator. As shown in Table 2, the responses generated by MMI is often short, and our responses for Daily dialog are long, thus influencing the Dist scores. Ungrammatical bigrams could cause Dist-2 to be high, as in MLE+sampling in DailyDialog.

\begin{table*}[ht]

\centering

(a) 

\scriptsize
\begin{tabular}{ll|ll}
  \toprule
  &{\bf Prompt $D_x$ (Light red)} && {\bf Prompt $D_x$ (Light green)}
  \\
  \cmidrule(lr){1-2}  \cmidrule(lr){3-4}
  & what instruments do you play ?&&what do you do for a living ? (Encoded by prompt encoder)\\
  \midrule
 
  & {\bf Response $D_y$ (Dark red)} && {\bf Response $D_y$ (Dark green) }
  \\
  \cmidrule(lr){1-2}  \cmidrule(lr){3-4}
  
  &i practice the piano every day .&&i work as an elementary teacher .
\\
  &i can play anything on my electric violin .&&i am an olympic gymnast 
\\
  &i am learning the guitar .&&aside from nursing , i work at a bar to pay for school .\\
  &i like the drums a lot&&i'm a janitor , but i also play music at night . you ?\\
  &i used to play clarinet .&&since i was fired i found a job in insurance .\\
  &i play trombone , alto sax , baritone , and trumpet . you ?&&i work part time as a bartender , but i don't drink any alcohol
\\
  &my parents taught me flute&&mechanical engineering is my day job .\\
  \midrule
 
  & {\bf Prompt $D_x$ (Light blue)} && {\bf Prompt $D_x$ (Light yellow)}
  \\
  \cmidrule(lr){1-2}  \cmidrule(lr){3-4}
    &what is your favorite color ?&&ya , are you a female ?\\
   \midrule
 
  & {\bf Response $D_y$ (Dark blue)} && {\bf Response $D_y$ (Dark yellow)}
  \\
  \cmidrule(lr){1-2}  \cmidrule(lr){3-4}
 
    &my favorite color is green and whats yours&&yes i am a woman . 

\\
  &i like red too , with a bit of yellow .  like a superhero !
&&not much to tell , i'm an average male . tell me about you .
\\
  &blue color makes me happy
&&female \\
  &mine is orange !
&&i am just a boy with a heart outside my body\\
  &i like rainbow colors , you ?
&&i am a 12 year old female
\\
  &red , blue , green , and yellow . i am thinking purple too
&&\\
  &strangely my favorite is grey !
&&\\
   \midrule
 
  & {\bf Response $D_y$ (Brown)} && {\bf Response $D_y$ (Black)}
  \\
  \cmidrule(lr){1-2}  \cmidrule(lr){3-4}
 & what do you do for a living ? (Encoded by response encoder)&&I don't know .\\
  
 \bottomrule
  
 \end{tabular}

\label{tab:responses}
\end{table*}

\begin{figure*}[h]
\centering
(b)

\includegraphics[scale=0.7]{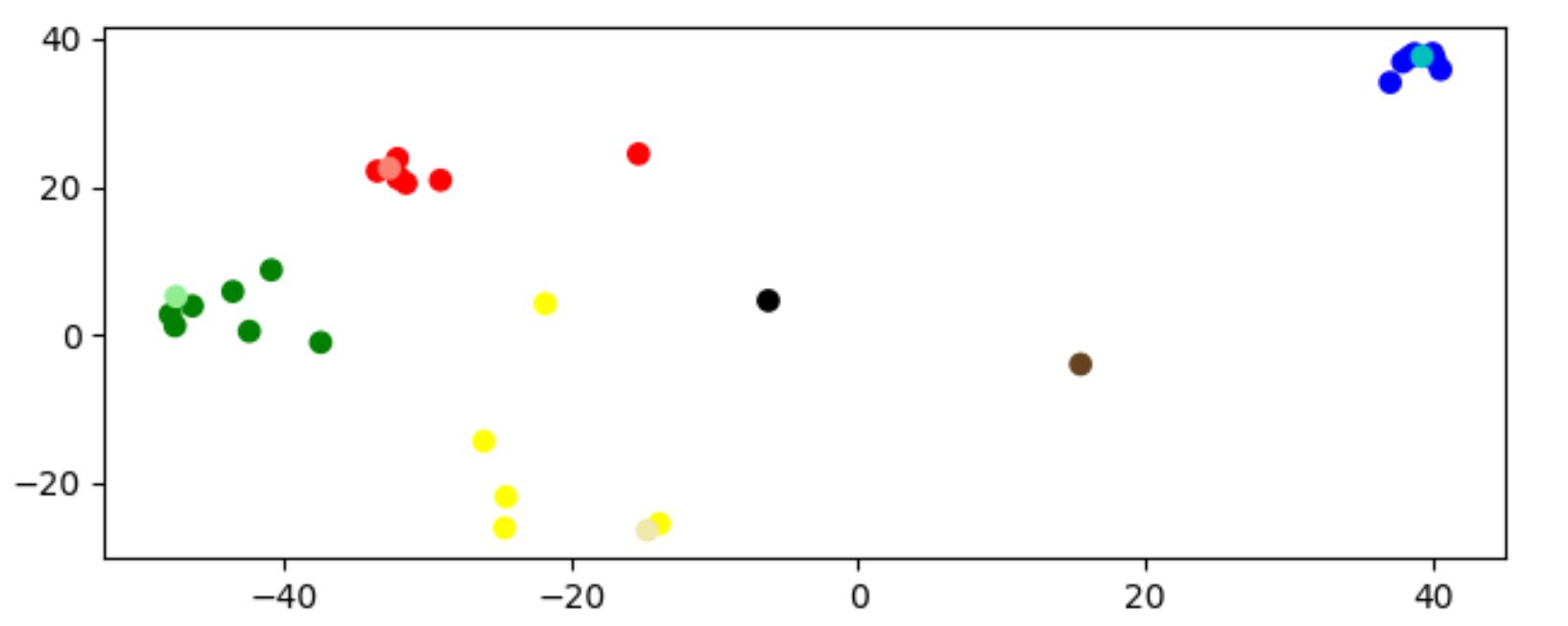}
\caption{t-SNE Visualization of the semantic latent space. The representations of the sentences in (a) are plotted in (b). Prompts ${D}_{x}$ and responses ${D}_{y}$ are encoded by separate encoders ${F}_{x}$ and ${F}_{y}$.  Multiple semantically related responses are close to each other and close to the corresponding prompt, while generic responses are far away. }
\label{fig:visualization}
\end{figure*}

\subsection{Ablation study}
We compare our full model with two variants and test the contribution of different parts in our model.
We use the PersonaChat dataset for this experiment.
The \textbf{w/o Uncorrelated part} model does not have the representation $Y_u$, the autoencoder reconstruction is solely based on $Y$, which also learns the CCA task.
In the \textbf{w/o Denoising} model, we do not replace random words with $\langle unk \rangle$ in the autoencoder input.

As shown in Table \ref{tab:results}, without the uncorrelated part, there is an obvious decrease in relevance, showing that allowing uncorrelated information is important for the learning the correlation between the prompt-response pairs. Without denoising, the generated sentences contain many grammatical errors. Since the sentences are obviously unacceptable by humans, we did not perform human evaluation. All automatic metrics also decreased, except Dist-2 is high because there are ungrammatical bigrams. This shows that denoising is critical for our model to generate grammatical responses. 

\subsection{Visualizing the semantic space}

In order to verify that the shared latent space successfully encodes semantic information, we visualize the representations of some sentences in Figure \ref{fig:visualization}. The dimension reduction is performed using t-SNE \cite{tsne} trained on \numprint{1000} prompt representations and \numprint{1000} response representations in the test set.

The light red point is the latent representation of the sentence {\em ``what instruments do you play ?''} encoded by the prompt encoder $F_x$. The seven dark red points are possible responses encoded by the response encoder $F_y$, such as {\em``i practice the piano every day .''}, {\em ``i am learning the guitar .''} Similarly, the light and dark blue, green, and yellow points show possible responses to three other questions. 

We can see that semantically related responses to the same question are clustered, showing that the latent space is indeed able to capture semantic information. The questions' representation is close to the cluster of their corresponding responses, demonstrating that our model has successfully learned from the collection of semantically similar possible responses. 

We also visualize {\em ``i don't know .''} in black, and the most frequent generic response of the Seq2Seq model trained on PersonaChat, {\em ``what do you do for a living ?''} in brown, using the response encoder $F_y$. Those generic responses are much farther away from the question than specific responses, thus they are unlikely to be generated. Note that the prompts and responses are encoded by separate encoders but plotted on the same space, so there are two points for ``what do you do for a living ?'', one as a prompt and the other as a response. 

\subsection{Grammaticality and comprehensibility}
Since generating text from a continuous space was previously found to produce grammatical errors \cite{gscs}, we show 500 PersonaChat responses, each to 3 crowdworkers to evaluate the grammaticality and comprehensibility of our model. We asked them to choose between the following options: About $11\%$ of sentences contain major grammatical errors that makes understanding the sentence difficult. $18\%$  contain minor errors that do not affect the understanding of the sentence. $71\%$ of the sentences are grammatically correct. This shows that most of the responses of our model are acceptable by humans, and comprehensibility is not a major problem for our latent space method.\footnote{MLE+beam search almost never makes grammatical errors as most responses are generic. While responses gets more informative and complicated, the issue of grammaticality becomes more probable.}

\section{Conclusion}
In this work, we pointed out that end-to-end cross entropy classification used in most previous methods is not able to integrate information from different semantically similar words responses, and designed a substitute method that is able to do so. Our method learns the pair relationship between prompts and responses as a regression task on a latent space, which is more suitable for the open-ended nature of this task. We performed ablation study to validate the components of our model. Human evaluation results concretely demonstrate that our latent space method significantly outperforms baselines using end-to-end cross entropy training, in terms of relevance and informativeness.

\bibliography{anthology,emnlp2020}
\bibliographystyle{acl_natbib}
\appendix
\clearpage

\end{document}